\documentclass[12pt]{iopart}
\makeatletter\if@twocolumn\PassOptionsToPackage{switch}{lineno}\else\fi\makeatother

\usepackage{tabulary,graphicx}
\usepackage[T1]{fontenc}
\usepackage[perpage]{footmisc}
    
\usepackage{lineno}
\usepackage[utf8]{inputenc} 
\usepackage{hyperref}       
\usepackage{url}            
\usepackage{booktabs}       
\usepackage{nicefrac}       
\usepackage{microtype}      

\usepackage[noend]{algpseudocode}
\usepackage{algorithmicx,algorithm}
\usepackage{caption}
\usepackage{tikz-cd}
\usepackage{subcaption}
\usepackage{wrapfig}
\usepackage{lipsum}
\usepackage{lmodern}
\usepackage{soul}
\soulregister\cite7 
\soulregister\citep7 
\soulregister\citet7 
\soulregister\ref7 
\soulregister\pageref7 

\newcommand\independent{\protect\mathpalette{\protect\independenT}{\perp}}
\def\independenT#1#2{\mathrel{\rlap{$#1#2$}\mkern2mu{#1#2}}}

\makeatletter
\expandafter\let\csname equation*\endcsname\relax
\expandafter\let\csname endequation*\endcsname\relax
\expandafter\let\csname subarray\endcsname\relax
\expandafter\let\csname endsubarray\endcsname\relax
\expandafter\let\csname substack\endcsname\relax
\expandafter\let\csname endsubstack\endcsname\relax
\def\[{\relax\ifmmode\@badmath\else
 \begin{trivlist}
 \@beginparpenalty\predisplaypenalty
 \@endparpenalty\postdisplaypenalty
 \item[]\leavevmode
 \hbox to\linewidth\bgroup$ \displaystyle
 \hskip\mathindent\bgroup\fi}
\def\]{\relax\ifmmode \egroup $\hfil \egroup \end{trivlist}\else \@badmath \fi}
\def\equation{\@beginparpenalty\predisplaypenalty
 \@endparpenalty\postdisplaypenalty
\refstepcounter{equation}\trivlist \item[]\leavevmode
 \hbox to\linewidth\bgroup $ \displaystyle
\hskip\mathindent}
\def\endequation{$\hfil \displaywidth\linewidth\@eqnnum\egroup \endtrivlist}
\@namedef{equation*}{\[}
\@namedef{endequation*}{\]}
\def\eqnarray{\stepcounter{equation}
   \def\@currentlabel{\p@equation\theequation}%
\global\@eqnswtrue
\global\@eqcnt\z@\tabskip\mathindent\let\\=\@eqncr
\abovedisplayskip\topsep\ifvmode\advance\abovedisplayskip\partopsep\fi
\belowdisplayskip\abovedisplayskip
\belowdisplayshortskip\abovedisplayskip
\abovedisplayshortskip\abovedisplayskip
$$\halign to
\linewidth\bgroup\@eqnsel$\displaystyle\tabskip\z@
 {##{}}$&\global\@eqcnt\@ne $\displaystyle{{}##{}}$\hfil
 &\global\@eqcnt\tw@ $\displaystyle{{}##}$\hfil
 \tabskip\@centering&\llap{##}\tabskip\z@\cr}
\def\endeqnarray{\@@eqncr\egroup
 \global\advance\c@equation\m@ne$$\global\@ignoretrue }
		
\usepackage{iopams,setstack}

\usepackage{url,multirow,morefloats,floatflt,cancel,tfrupee}
\makeatletter

\AtBeginDocument{\@ifpackageloaded{textcomp}{}{\usepackage{textcomp}}}
\makeatother
\usepackage{colortbl}
\usepackage{xcolor}
\usepackage{pifont}
\usepackage[nointegrals]{wasysym}
\makeatletter

\def\mcWidth#1{\csname TY@F#1\endcsname+\tabcolsep}

\def\cAlignHack{\rightskip\@flushglue\leftskip\@flushglue\parindent\z@\parfillskip\z@skip}
\def\rAlignHack{\rightskip\z@skip\leftskip\@flushglue \parindent\z@\parfillskip\z@skip}

\@ifundefined{etal}{}{}

\usepackage{ifxetex}
\ifxetex\else\if@twocolumn\@ifpackageloaded{stfloats}{}{\usepackage{dblfloatfix}}\fi\fi

\AtBeginDocument{
\expandafter\ifx\csname eqalign\endcsname\relax
\def\eqalign#1{\null\vcenter{\def\\{\cr}\openup\jot\m@th
  \ialign{\strut$\displaystyle{##}$\hfil&$\displaystyle{{}##}$\hfil
      \crcr#1\crcr}}\,}
\fi
}

\AtBeginDocument{%
  \@ifpackageloaded{endfloat}%
   {\renewcommand\efloat@iwrite[1]{\immediate\expandafter\protected@write\csname efloat@post#1\endcsname{}}}{\newif\ifefloat@tables}%
}%

\def\BreakURLText#1{\@tfor\brk@tempa:=#1\do{\brk@tempa\hskip0pt}}
\let\lt=<
\let\gt=>
\def\processVert{\ifmmode|\else\textbar\fi}

\@ifundefined{subparagraph}{
\def\subparagraph{\@startsection{paragraph}{5}{2\parindent}{0ex plus 0.1ex minus 0.1ex}%
{0ex}{\normalfont\small\itshape}}%
}{}

\newcommand\role[1]{\unskip}
\newcommand\aucollab[1]{\unskip}
  
\@ifundefined{tsGraphicsScaleX}{\gdef\tsGraphicsScaleX{1}}{}
\@ifundefined{tsGraphicsScaleY}{\gdef\tsGraphicsScaleY{.9}}{}
\def\checkGraphicsWidth{\ifdim\Gin@nat@width>\linewidth
	\tsGraphicsScaleX\linewidth\else\Gin@nat@width\fi}

\def\checkGraphicsHeight{\ifdim\Gin@nat@height>.9\textheight
	\tsGraphicsScaleY\textheight\else\Gin@nat@height\fi}

\def\fixFloatSize#1{}
\let\ts@includegraphics\includegraphics

\def\inlinegraphic[#1]#2{{\edef\@tempa{#1}\edef\baseline@shift{\ifx\@tempa\@empty0\else#1\fi}\edef\tempZ{\the\numexpr(\numexpr(\baseline@shift*\f@size/100))}\protect\raisebox{\tempZ pt}{\ts@includegraphics{#2}}}}

\AtBeginDocument{\def\includegraphics{\@ifnextchar[{\ts@includegraphics}{\ts@includegraphics[width=\checkGraphicsWidth,height=\checkGraphicsHeight,keepaspectratio]}}}

\DeclareMathAlphabet{\mathpzc}{OT1}{pzc}{m}{it}

\def\URL#1#2{\@ifundefined{href}{#2}{\href{#1}{#2}}}

\def\UrlOrds{\do\*\do\-\do\~\do\'\do\"\do\-}%
\g@addto@macro{\UrlBreaks}{\UrlOrds}

\edef\fntEncoding{\f@encoding}

\makeatother

\newif\ifmultipleabstract\multipleabstractfalse%
\usepackage[numbers,square,sort&compress]{natbib}

\usepackage{float}

\begin{document}


\title[]{Physics-Informed Generative Neural Network: An Application to Troposphere Temperature Prediction}

 \author{Zhihao Chen$^{a,+}$, Jie~Gao$^{b,+,*}$, Weikai Wang$^b$, Zheng Yan$^{b,*}$}
 \address{$^{a}$East China Regional Air Traffic Management Bureau, CAAC, Shanghai\unskip, China}
 \address{$^{b}$AI Lab\unskip, Shanghai Em-Data Technology Co. Ltd., Shanghai\unskip, China}
 \address{$^{*}$Corresponding author}
  \address{$^{+}$The authors contributed equally.}
 \ead{chen\_zhihao@126.com}
 \ead{\{gaojie1,wangweikai,yanzheng\}@em-data.com.cn}

\begin{abstract}
The troposphere is one of the atmospheric layers where most weather phenomena occur. Temperature variations in the troposphere, especially at 500hPa, a typical level of the middle troposphere, are significant indicators of future weather changes. Numerical Weather Prediction (NWP) is effective for temperature prediction, but its computational complexity hinders a timely response. This paper proposes a novel temperature prediction approach in framework of physics-informed deep learning. The new model, called PGnet, builds upon a generative neural network with a mask matrix. The mask is designed to distinguish the low-quality predicted regions generated by the first physical stage. The generative neural network takes the mask as prior for the second-stage refined predictions. A mask-loss and a jump pattern strategy are developed to train the generative neural network without accumulating errors during making time-series predictions. Experiments on ERA5 demonstrate that PGnet can generate more refined temperature predictions than the state-of-the-art.  

\end{abstract}\def\keywordstitle{Keywords}

\vspace{2pc}\noindent\textit{Keywords: }{physics-informed deep learning, temperature prediction, generative neural network, video prediction}
    
\section{Introduction}
Weather forecasting plays a significant role in various fields \cite{Kornhuber-ERL, Thornton-ERL}, including civil aviation, society-level emergency, and agriculture activities. And, with the development of the economy and society, more accurate and timely weather forecasting is becoming an urgent demand. The mainstream weather forecast methods are physical process-based methods like numerical weather prediction (NWP) \cite{02lynch2008origins,stanger2019optimising}. Since, from the physical process perspective, weather forecasting is a sustained phenomenon marked by gradual changes through a series of states occurring in the physical world \cite{02lynch2008origins}. Thus, it can be solved by physical dynamics \cite{01sekula2019prediction,30brunton2016discovering}. NWP leverages the integrated data of different weather services on coarse-grained resolution grids covering wide geographical areas and describes several meteorological variables such as temperature, humidity, geopotential height, wind components, Etc. These meteorological variables define the predicted atmospheric pattern for a given forecast period. However, NWP does not provide timely responses due to its high computation complexity.
Recently, with the significant achievements of deep learning in various areas, researchers have also set out exploring the methodology of deep neural networks for this time-series forecasting task \cite{10ham2019deep,09roscher2020explainable,Yan2020,Feng2021,Feng2021b} . Actually, these deep learning-based techniques commonly regard this task as a video prediction task \cite{24xingjian2015convolutional,wang2018predrnn++,jin2020exploring}. They extract the spatial and temporal contexts of historical meteorological observations to train specific forecast models. Compared with NWP, these specific models can provide more timely prediction with powerful specialized hardware such as GPUs and TPUs, which is more effective than CPUs.

One main branch of the deep learning methodology is a spatiotemporal neural network that relies on 3D convolutions and recurrent networks to capture high-level spatial and temporal features, hoping to extrapolate future observations \cite{24xingjian2015convolutional,25shi2017deep,pang2019deep}. For the precipitation nowcasting task, it utilizes the past radar echo chart sequences to predict future radar echo maps. Shi et al. \cite{24xingjian2015convolutional} proposed the Convolutional Long Short-Term Memory(ConvLSTM) model by extending the conventional LSTM \cite{lstmgreff2016lstm} through adding the convolutional operations into both input-to-state and state-to-state. Later they introduced the trajectory GRU model \cite{25shi2017deep} that can actively learn the location variant structure for recurrent connections. Bo Pang et al.\cite{pang2019deep} raised a new ST-LSTM with a new deep recurrent structure named DeepRNN for spatiotemporal prediction. But, these spatiotemporal network models may suffer from blurry effects, especially in long-term prediction.

The other active branch is the flow-based approaches that seek pixels' moving trails through the calculated motion filed by deep neural networks \cite{liang2017dual, 38wang2018video,ilg2017flownet}. Motion-based approaches \cite{liu2017video,liang2017dual,kwon2019predicting, 38wang2018video} can generate sharp images but fail when the conflict pixels generated. Some works introduces GAN \cite{goodfellow2014generative,zhu2017unpaired},  which simultaneously trains discriminator and generator networks to generate  more realistic images, into the prediction task. Dual Motion GAN \cite{liang2017dual, 38wang2018video} synthesis images by combines pixel-wise motion \cite{ilg2017flownet} and generative adversarial network. However, prior physical knowledge is all ignored by these approaches. 

How to incorporate the advantages of both deep learning approaches and the known physical process simultaneously still remains challenging\cite{han2018solving,03de2019deep,06onfared2009new,raissi2019physics}. To improve prediction models by exploiting prior physical knowledge, early some scientists focused on governing physical equations underlying a dynamical system simply from data measurement \cite{31berg2019data,30brunton2016discovering}. and some others used neural networks to approximate the solution of PDEs \cite{32raissi2018deep,raissi2019physics,33seo2019differentiable}. Maziar \cite{raissi2019physics} introduced physical informed neural networks that trained to solve supervised learning tasks. Sung Yong and Yan Liu \cite{33seo2019differentiable, seo2019physics} raised differentiable physics-informed graph networks (DPGN) that incorporate implicit physics knowledge by informing it to latent space. Recently, some works \cite{03de2019deep,han2018solving,wang2020towards,34weinan2017proposal} are dedicated to specific PDEs like advection-diffusion, Euler, Navier–Stokes. Bezenac constructed a deep learning framework that benefited from physical processes and presented the effectiveness of prior scientific knowledge of the convection-diffusion equation to improve its performance. PDE-Net \cite{long2018pde,36long2019pde} discretizes a broad class of PDEs by approximating partial derivatives with convolution neural network. Vincent \cite{37guen2020disentangling} proposed PhyDNet, a two-branch deep architecture, which explicitly disentangles PDE dynamics from unknown complementary information. Although these works leverage physical knowledge, they either ignored the unknown physics laws or ignored mitigate conflicts in the generated results \cite{27gao2019disentangling}. What is worse, that the conflict phenomenon frequently happens in image generatIn this paper, we propose a two-stage model called PGnet  that benefits from the both advantages: 1). the excellent numerical fitting ability of deep learning approaches that is commonly with the auxiliary of high-performance computing units. 2). the higher accuracy  of physical process. To our best knowledge, no previous work has utilized physical processes and generative networks simultaneously. This phenomenon will be discussed in detail in section \ref{sec:loss}.

In this paper, we propose a two-stage model called PGnet  that benefits from the both advantages: 1). the excellent numerical fitting ability of deep learning approaches that is commonly with the auxiliary of high-performance computing units. 2). the higher accuracy  of physical process. The first-stage is the physic-informed method that generates propagated images, which could involve some conflict pixels and blurry regions. The second-stage is the generative neural model that  can calibrate and improve the first-stage outputs. To our best knowledge, none of any works has utilized physical processes and generative networks simultaneously. Furthermore, a MASK matrix  is designed to distinguish these conflict contents, then the it will be feed into the second-stage. We also devise a mask-loss to adjust the contributions of these two stages dynamically, and put froward an evolution method with a jump pattern strategy for time-series multi-frame prediction.

Our contributions are summarized below: 

\begin{enumerate}
 \item We proposed a two-stage model called PGnet that incorporates a physic-informed method and an image synthesis neural network for troposphere temperature prediction.
 \item We put forward an evolution method with a jump pattern strategy for time-series multi-frame prediction.
    \item We evaluate PGnet on the troposphere temperature dataset and achieve more refined results than the previous methods.
\end{enumerate}

This paper is organized as follows: . Section \ref{sec:background} describes the background of convection-diffusion equation. Section \ref{sec:method} elaborate PGnet. Section \ref{sec:experiments} describes the training details and the performance evaluation. Conclusion and discussion are shown in Section \ref{sec:conclusion}.

This paper proposed a two-stage model PGnet that jointly takes known physical laws governed elements and unknown laws governed or conflict happened elements into account. The physics processes help to predict known physical laws governed elements. The generator network helps to capture unknown laws and to synthesis conflict elements.

\section{Background}
\label{sec:background}
Partial differential equations (PDEs) play a prominent role in many disciplines of science and engineering. It describes a wide variety of physical phenomena such as sound, heat, fluid dynamics. A specific PDE equation can describe how some physical phenomena transfer into a physical system, such as  physical particles, energy(temperature). And it is called the convection-diffusion equation that integrates the diffusion process and the convection process. The traditional method forecasts temperature by solving the convection-diffusion equation using numerical method \cite{03de2019deep}.
The convection-diffusion equation is illustrated in equation $1$. Let $w$ be the vector velocity field of the flow with two components $(u, v)$, velocities along with $x$ and $y$ directions, $T$ be the temperature, the governing equations for this physical system are :

$$\frac{\partial T}{\partial t}+(w.\nabla) T=D \nabla^{2} T  \quad ,\quad \nabla \cdot \boldsymbol{w}=0  \eqno{(1)} $$
where $\nabla$ denotes the gradient operator, $\nabla^{2}$ 
denotes the Laplacian operator, and $D$ is the diffusion coefficient.
According to theorem 1 of \cite{03de2019deep}, for  the initial condition $T_{0}$, there exists a unique global solution $T(x, t)$ to the convection-diffusion equation 1 :


\begin{center}
$$
\setlength{\abovedisplayskip}{-10pt}
\setlength{\belowdisplayskip}{0pt} 
T(x, t)=\int_{\mathbb{R}^{2}} k(x-w, y) T_{0}(y) d y     \eqno{(2)}$$
\end{center}
where $k(u, v)=\frac{1}{4 \pi D t} e^{-\frac{1}{4 D t}\|u-v\|^{2}}$ is the kernel. Provide the motion $w$ and the diffusion coefficient $D$. It states that for any timestamp and location, temperature $T(x, t)$ can be calculated by a convolution operation between an initial condition $T_{0}$ and a Gaussian probability density function. In other words, if the troposphere temperature underlying advection mechanisms were known, future troposphere temperature could be predicted from previous ones. Unfortunately, neither the initial conditions, the motion vector, nor the diffusion coefficient is known. They have to be estimated from data.  Using the same method in \cite{03de2019deep}, in the first-stage of PGnet, we learn to predict a motion field analog to the $w$ in equation $3$ to generate intermediate physical processes results. 
Discretizing equation $2$. by replacing the integral with a sum, and setting temperature field $T_{i}$ as the initial condition, we can calculate the future temperature field  $T_{i+1}$ based on the motion field estimate $\hat{w}$. 

\begin{center}
$$
\setlength{\abovedisplayskip}{-10pt}
\setlength{\belowdisplayskip}{0pt} 
\hat{T}_{i+1}(x)=\sum_{y \in \Omega} k(x-\hat{w}(x), y) T_{i}(y) \eqno{(3)}$$
\end{center}
As seen by the relation with the solution of the advection-diffusion equation, we use the warping mechanism in \cite{03de2019deep} that clearly adapted to the modeling of phenomena governed by the advection-diffusion equation. Troposphere temperature forecasting is a particular case.

\section{Method}
\label{sec:method}

We view the forecasting of temperature as a video prediction task, giving a series of historical N frames $X^{N}=\left\{T_{1}, T_{2}, \ldots, T_{N}\right\}$ and predicting the future K frames $X^{K}=\left\{T_{N+1}, T_{N+2}, \ldots, T_{N+K}\right\}$. Each frame can be perceived as a local area temperature field. 

\begin{figure}[!ht]
    
  \centering
  
  \begin{subfigure}[t]{0.5\linewidth}
    \parbox[][2.5cm][c]{\linewidth}{
      \centering
          \includegraphics[width=0.9\linewidth]{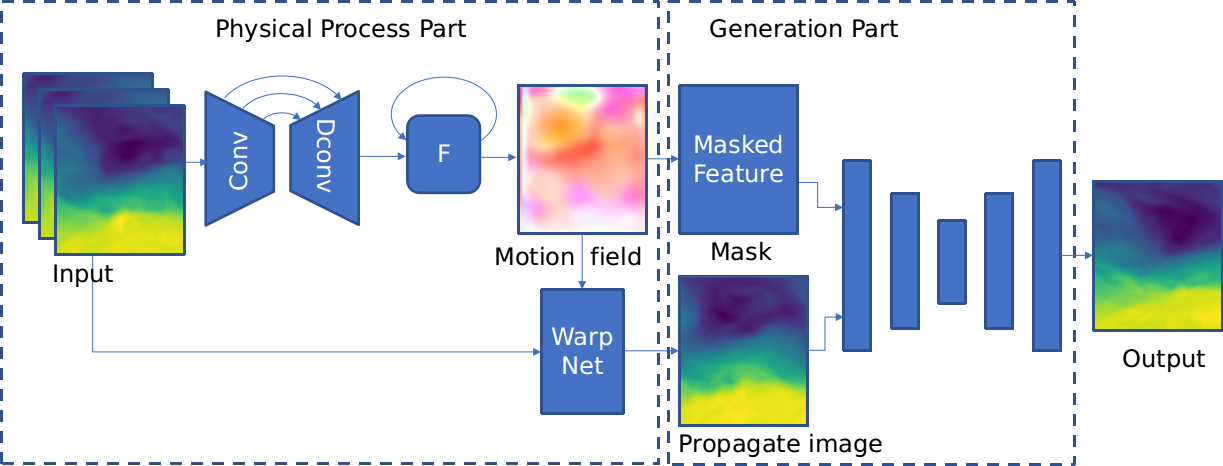}}
      \caption{}
  \end{subfigure}
  \begin{subfigure}[t]{0.45\linewidth}
    \parbox[][2.5cm][c]{\linewidth}{
      \centering
          \includegraphics[width=0.8\linewidth]{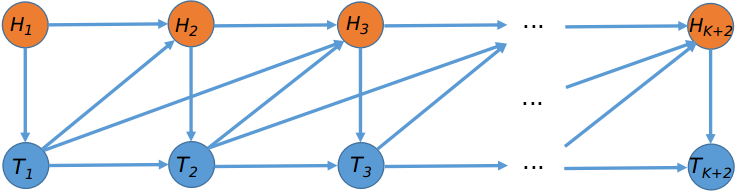}}
      \caption{}
  \end{subfigure}

  \caption{The overall architecture is shown in (a). The motion field is generated from input images with skip connected convolutional neural networks. F is a motion field evolution function that learns time-dependent dynamic characteristics of the motion field. The last frame of inputs is warped by the motion field to predict the propagate image. The generation part is also a CNN with encoder-decoder architecture that concatenates propagate image and the mask feature as input to generate the future frame. (b) shows the probabilistic graphical model representation of the proposed motion evolution method in which the evolution order $M=1$, the number of inputs frames $N=2$, and predicts $K$ frames.}
\vspace{-3mm}
\label{fig:main}
\end{figure}

As shown in figure \ref{fig:main}, we design a two-stage model PGnet to disentangles physical process propagation from physical-agnostic generation. The first physical processing stage propagates the pixels constrained by convection-diffusion equations using the method described in \ref{sec:background}. The first-stage generates a mask matrix and the propagated image that contain conflicting pixels.
We define conflicting pixels, including boundary-affected pixels and collision pixels. As we treat the flow field as images, the images' boundary pixels propagated from previous images' outer areas are called boundary affected pixels. There are two cases for collision pixels. Case one: there is no pixel propagate to this coordinate, but boundary conditions do not cause it. Case two: there are at least two pixels propagate to this coordinate.
A mask matrix that distinguishes the conflicting pixels is used. Please refer to section \ref{sec:loss} for the detailed description of mask matrix.
The evolution method with a jump pattern strategy used in the first stage will be described in sections \ref{sec:jp} and \ref{sec:me}.

Moreover, the second stage network's purpose is to generate the unreliable pixels indicated in the mask matrix. Pixels are not isolated. There are many cases\cite{bertalmio2000image,barnes2009patchmatch,zhang2018context} where appropriate spatial contexts must be retrieved. Spatial context encoders\cite{pathak2016context} query learned dataset priors with an exposed appearance searching for missing patches. We use the generation network stage to encode the propagated image's spatial context into a latent space. By employing the decoder network, we aim to generate the temperature field, wildly conflicting pixels from the latent space.


    
\subsection{Model Components}
\begin{algorithm}[t]
  \caption{Jump Pattern Algorithm} 
  \hspace*{0.02in} {\bf Input:} 
  Require data $T_{1:N}$, number of predict frames K\\ 
  \hspace*{0.02in} {\bf Output:} 
  Predicted images $\hat{T}_{N+1:N+K}$ 
  \begin{algorithmic}[1]
  \State Get the last input frame $T_{N}$
  \For{i = 0 to K} 
      \State Predicte motion field $dW$ with physical process network
      \If{i==0}
        \State $W = dW$
        \State $dW_{temp} = dW$
    \Else
        \State $dW_{temp} = F(dW_{temp},dW)$
        \State $W = dW_{temp}+warp(W,dW)$ 
       \EndIf
          \State $\hat{\hat{T}}_{N+i} = warp(T_{N},W)$ 
          \State Compute mask $M$ with motion field $W$
          \State Predicte image $\hat{T}_{N+i}$ with $\hat{\hat{T}}_{N+i}$ and $M$ using generator network 
  \EndFor
  \State \Return $\hat{T}_{N+1:N+K}$  
\end{algorithmic}
\end{algorithm}
For the physical process stage of PGnet, Conv-Deconv Network with skip connections is first used to produce the interval motion field. The final motion field is computed by motion evolution function F which will be explained in \ref{sec:me}. A warp scheme described in section \ref{sec:background} that can apply the advection-diffusion constraint is employed to predict the propagate image. Different from \cite{03de2019deep}, the warped image that contains conflicting pixels are served as the final output, the propagate image is utilized as an intermediate feature map in our design. 
For the generation network, a convolutional encoder-decoder network with symmetric skip connections is used.
First, a mask is computed by the motion field to distinguish conflicting pixels. 
The output propagate image is channel-wise concatenated with mask feature and fed into the generation network. In order to improve accuracy and generate realistic images, 
an evolution method with a jump pattern strategy is put forward for multi-frame prediction and will be discussed in the below subsection.

\subsection{Jump Pattern Algorithm}
\label{sec:jp}

Most of the approaches (ConvLstm\cite{24xingjian2015convolutional}, DeepRNN\cite{pang2019deep}, MCnet\cite{mcnetvillegas2017decomposing}, Traj-GRU\cite{25shi2017deep}, Vid2Vid\cite{38wang2018video}, etc\cite{nemgreff2017neural}\cite{zhou2020deep}) implement the next-step prediction objective by replacing $T_{n+1}$ with $\hat{T}_{n+1}$ and evaluated at each time-step. These approaches make the hypothesis that front predictions are ground truth for later prediction. However, this hypothesis is extremely strict and leads to the blurry unrealistic image and decrease accuracy, especially for the later frames. 

To solve the problem mentioned above and reduce the error accumulation effect, we
introduce a jump pattern algorithm shown in algorithm, where $T_{i}$ represents the $i$th frame,
$dW_{temp}$ represents the cache interval motion field, $dW$ represents the current interval motion field. 
$W$ represents the final motion field, $M$ represents the computed mask. $warp(A, B)$ represents the warp network that deformation the input tensor $A$ by the motion field $B$. 
It should be noted that $warp(A, B)$ is also used to warp the motion field in our algorithm which tracks pixels over time steps that distinguish it from previous works. $F(.)$ is motion evolution function. Different from the original method in the Journal of Statistical Mechanics \cite{03de2019deep} that the next frame is distorted from the previous frame, the frames of all steps are warped from the last frame of input images in our method. As a result, the times of warp and interpolation are reduced to one for each image and the accumulation error is restrained. We verify the effectiveness of this algorithm in the experiment section.

\subsection{Motion Evolution}
\label{sec:me}
The natural spatiotemporal processes like troposphere temperature can be highly non-stationary in many ways. From Cramèr's Decomposition \cite{cramer1961some}, any non-stationary process can be decomposed into deterministic, time-variant polynomials, plus a zero-mean stochastic term. 
We formally develop a probabilistic graphical model to learn time-variant features by introducing a motion evolution order $M$ which represents the current motion is dependent on previous $M$ motions. 
Generally, the motion field can be seen as a hidden state of different steps. A specific graph G is illustrated in fig:\ref{fig:main} (b) with evolution order $M=1$  inputs $N=2$ images and predicts $K$ frames without jump pattern. 
Where $H$ represents the hidden state of different prediction steps (e.g., interval motion field).
The structure of the graph G induces the following conditional independence relationships. 
\begin{center}
    \setlength{\abovedisplayskip}{-10pt}
  \setlength{\belowdisplayskip}{5pt} 
  $$T_{i+1} \independent T_{\leq i-1} | H_{i+1},T_{i}   \quad       \forall i=2,...,K+1 $$  
  $$H_{i} \independent T_{\leq i-3} | H_{i-1},T_{i-1},T_{i-2}   \quad       \forall i=4,...,K+2 $$  
\end{center} 
Generally, for evolution order $M$, inputs $N$ images and predict $K$ frames, we have the following conditional independence relationships.
That is, the future observation is independent of past observations gave the current observation and future hidden state, and the current $i$th hidden state is independent of $i-N$ past observations given the previous $M$ hidden state and previous $N$ observation.
By the following equation, we can model the joint distribution of $T_{1:N+K}$ and $H_{1:N+K}$ by using the motion evolution method.
  \begin{center}
    $$
    \setlength{\abovedisplayskip}{-10pt}
    \setlength{\belowdisplayskip}{5pt} 
    P(T_{1:N+K},H_{1:N+K}) = \prod_{i = max(M,N)+1}^{N+K}P(H_{i}|H_{i-M:i-1},T_{i-N:i-1})P(T_{i}|H_{i},T_{i-1})    \eqno{(4)}$$    
  \end{center} 
  In our implementation $dW_{temp}=P(H_{i}|H_{i-M:i-1},T_{i-N:i-1})$ and can be calculated by $F(.)$. $P(T_{i}|H_{i},T_{i-1})$ is achieved by $warpnet$. It should be pointed out that $F(.)$ could have different forms, and two kinds of implement function $F(.)$ is provided in our study to learn the time-varying features directly on the predicted motion fields other than original frames.
  CNN based function $F(.)$: $dW_{temp} = CNN(dW,warp(dW_{temp},dW)) $.
  Momentum based function $F(.)$: $dW_{temp} = (1-\beta)*dW + \beta*dW_{temp}$.
  As Lucas–Kanade method \cite{lucas1981iterative} assumes, the optical flow is essentially constant in a local neighborhood. We argue that such locality can be learned by CNN based function $F(.)$. And the previous $M$ motion fields are concatenated with the current motion field and input into the convolutional network.
  Besides, inspired by inertia, the motion vector generally does not vary dramatically. And humans can track the trajectory of objects with the latent notion of inertia. For our task, the motion field of different prediction steps should maintain stability and continuity. In order to obtain the beneficial effect of inertia, momentum-based function $F(.)$ is proposed ($M=1$ in this case). Where $\beta \in [0, 1)$ is momentum coefficient. 
  
  \subsection{Mask-Loss}
\label{sec:loss}
  Compared to deep learning methods, boundary conditions are often critical in the physical process methods like NWP. For a flow field image, most deep learning methods treat the pixels propagate within the image boundaries. This assumption is limited for the small region because boundary pixels are often propagated from outer areas. Besides, pixels also have occlusion problems and the "ghosting" effect\cite{27gao2019disentangling} when regions in the target frame to which extrapolated optical flow have no projection. Motivated by these ideas, a mask that can distinguish conflicting pixels, including boundary conditions affected pixels and collision pixels, is introduced. Furthermore, a mask-loss is designed to dynamically adjust the contributions of the physics process and generation network. 
  
  An energy-based method to compute the occlusion map from the optical flow is mentioned in \cite{27gao2019disentangling,finn2016unsupervised,hao2018controllable}. Differed from \cite{27gao2019disentangling}, the mask in our method is achieved by the motion field that generated from the physical process stage of PGnet. Besides, the boundary conditions are also considered. The pixel density can be viewed as an energy map. In order to distinguish pixels influenced by border inspection conditions, a specific padding value(zero in our study) is used in the warp scheme\cite{03de2019deep}. Consequently, the pixels propagate from the outer area can be identified. We aim to compute a mask matrix that distinguishes conflicting pixels, including boundary-affected pixels and collision pixels. Similar to the method in \cite{27gao2019disentangling}.
  We initialize the energy field of the first frame to be a matrix filled with ones, denoted as $E_{x,y}^{1}$.
  Given a motion field, for each pixel in the first frame, the energy unit on each coordinate will be computed by its corresponded coordinates in the second frame using the equation $4$. Given the motion field and we get a new energy field $E_{x,y}^{2}$. We consider two special cases for each coordinate $(x, y)$ in the second frame:
  1. If $E_{i,j}^{2} = 0$.  there is no pixel moving to this coordinate(mostly the boundary affected pixels padded by warp net) 
  2. If $E_{i,j}^{2} \geq 2$. there are at least two pixels in the first frame compete for the same location, which suggests it occluded and not trustable. 
  The mask are generated by $E_{i,j}^{2}$ values.  

  $$
  M_{i,j}\;=\;\left\{\begin{array}{l}0 \quad \forall E_{i,j}^{2} = 0 \quad  or \quad  E_{i,j}^{2} \geq 2\\ 1 \quad       \forall E_{i,j}^{2} = others 
  \end{array}\right. \eqno{(5)}$$

To dynamically adjust the contributions of physics process and generation network a masked loss is designed. The masked loss and overall loss is formulated below.
\begin{center}
$$
\setlength{\abovedisplayskip}{-10pt}
\setlength{\belowdisplayskip}{5pt} 
L_{mask}=\alpha * M * MSE(T ,\hat{T} ) + (1-\alpha)*(1-M)*MSE(T,\hat{T})            \eqno{(6)}$$
$$
L = \lambda_{\mathrm{lp}}*L_{mask}+\lambda_{\mathrm{div}}\left(\nabla \cdot w_{t}(x)\right)^{2} + \lambda_{smoth}\left\|(\nabla w_{t}(x)\right\|^{2}
$$
\end{center}
Where $N = W*H$ is the number of pixels. $W$ and $H$ are the width and height of the image. The parameter $\alpha$  is used to balance the weights of different parts.
Different from the magnitude regularization in \cite{03de2019deep}, which causes unstable of the model because of its negative coefficient.  We just additionally apply divergence loss and smoothness loss.

\section{Experiments}
\label{sec:experiments}

\subsection{Dataset}
\label{esc:dataset}
The dataset used in our experiment comes from ERA5\cite{hersbach2020era5}, a climate reanalysis dataset from ECMWF (European Centre for Medium-Range Weather Forecasts)\cite{molteni1996ecmwf}.
The ERA5 dataset offers various atmospheric, land-surface, and oceanic variables in a spatial resolution of 31 km, a time resolution of 1 hour, and 137 vertical levels. In this experiment, temperature data of 500 hPa level is used and sampled at 6-hour intervals within 1980-01-01 00:00 UTC to 2018-01-01 00:00 UTC. The sampled data of the adjacent three days form a sequence of 12 images in time order. Then we got 4510 image sequences in sum; 85\% is used to training 5\% for validation and 10\% for testing. The latitude and longitude range of the dataset is $10.5^{\circ} N$ to $74^{\circ}N$ and $72^{\circ}E$ to $135.5^{\circ}E$, with a spatial resolution of 0.5 degrees (about 50 km). 

\subsection{Baseline Comparison}

We evaluated our approach on the ERA5 dataset and compared it with several baselines. The model with 4 input frames is forecasting images on a horizon of 8 and evaluated with the mean square error (MSE), peak signal-to-noise ratio (PSNR), structural similarity (SSIM) metrics and pattern correlation (CORR).  
Besides, MSE scores are also given by different steps. The hyperparameters are tuned using the validation set. Models are trained on V100 NVIDIA GPUs. 
For measuring the effectiveness of the jump pattern algorithm and the motion evolution method, a progressive ablation experiment is designed with motion evolution order $M=1$. Higher-order evolution is an interesting area for future research.
Two variants of our model with different forms of function $F(.)$ are evaluated. 
For CNN based function $F(.)$, two $3*3$ convolutional layer and one $1*1$ convolutional layer\cite{Zhang_2018_CVPR} is stacked to capture spatial-temporal correlation of the motion field.
\begin{table}[!htbp]
      \centering
      \caption{Quantitative Result of the Ablation Experiments. Generation Network is represented as G and using the generation part of the proposed model. Jump pattern is represented as J and using the jump pattern algorithm. Motion evolution is represented as M and using the momentum method of function F(.). MSE, SSIM, PSNR are evaluated metric scores.}
      \begin{tabular}{p{1.6cm}p{1.6cm}p{1.6cm}p{1.75cm}p{1.75cm}p{1.75cm}p{1.75cm}}
      \toprule  
      G&J&M& MSE$\downarrow$ &SSIM$\uparrow$ & PSNR$\uparrow$\tabularnewline
      \midrule  
      &  &  & 12.210 & 0.820 & 28.884 \tabularnewline
      \checkmark&  &  & 9.775 & 0.889 & 31.199  \tabularnewline
      \checkmark &\checkmark& & 9.598 & 0.886 & 31.230  \tabularnewline
      \checkmark&\checkmark & \checkmark & \textbf{8.877} & \textbf{0.894} & \textbf{31.987} \tabularnewline
      \bottomrule 
      \end{tabular}
      \label{tb:ablation}
    \end{table}
Momentum based function $F(.)$ is designed as PGnet-Momentum that concatenates motion field with propagating image and the mask feature as the input of the generator network. The hyperparameter $\alpha$ was set to 0.9 by cross-validation.
Our model can be viewed as an upgraded version of the model proposed in \cite{03de2019deep}, a physically-constrained advection-diffusion flow model. Compared with \cite{03de2019deep}, our method not only applies physical constraints but also uses more deep learning methods like the generator network and motion evolution. We also introduce two purely deep learning models ConvLSTM and DeepRNN, as our baselines. ConvLSTM uses convolutional transitions in the inner LSTM module, and DeepRNN \cite{pang2019deep} can stack RNN deep effectively.

\begin{figure}[h!]
  \centering
  \begin{subfigure}[t]{0.485\linewidth}
      \centering
      \includegraphics[width=1.0\linewidth]{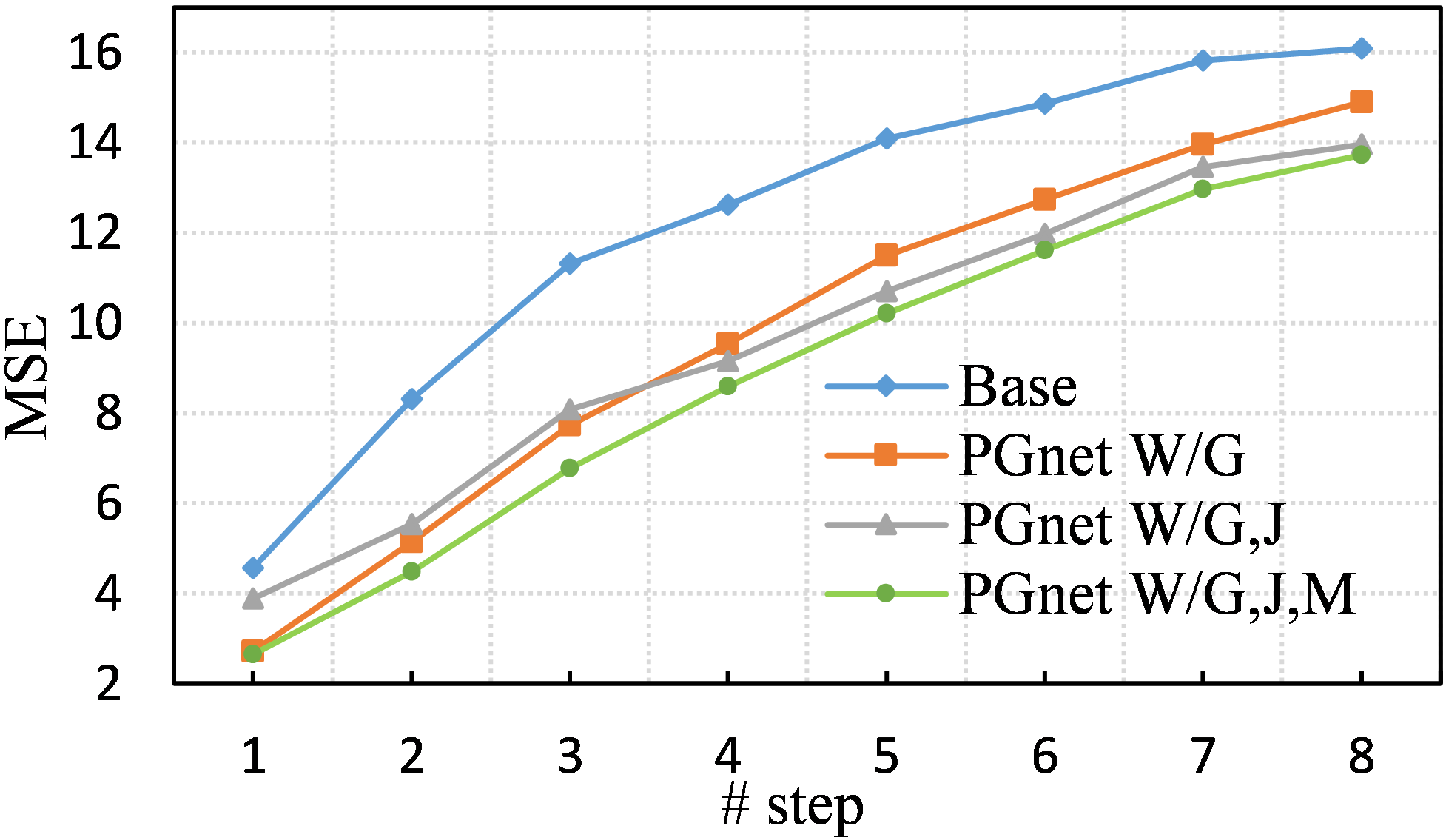}\vspace{0.08cm}
  \end{subfigure}
  \hspace{-0.06cm}
  \begin{subfigure}[t]{0.505\linewidth}
      \centering
          \includegraphics[width=1.0\linewidth]{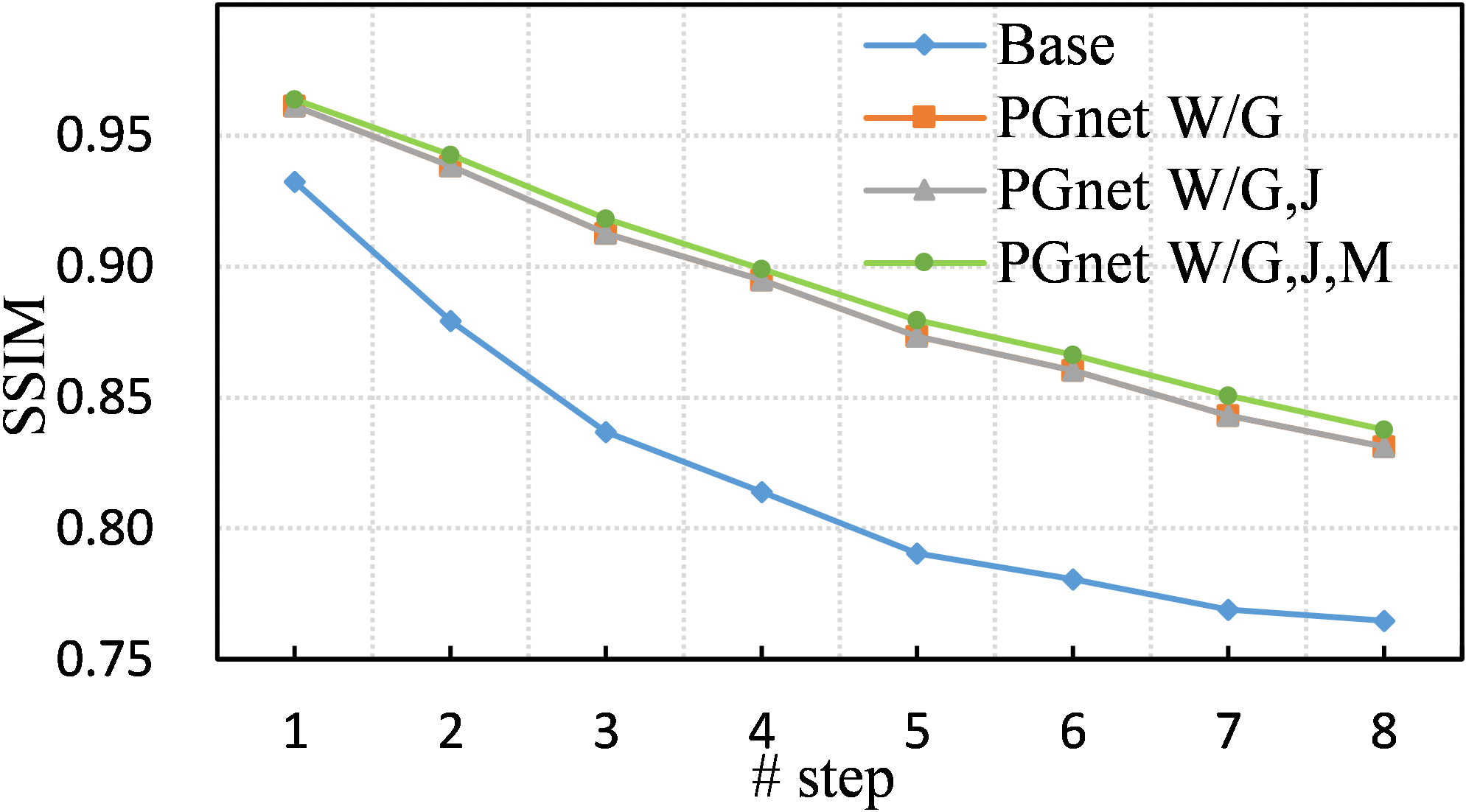}\vspace{0.08cm}
  \end{subfigure}
  \caption{ The averaged metric scores for different prediction steps.  We averaged the results of the testing data described in \ref{esc:dataset}.  The base model is CDNN with regularization(Emmanuel et al.(2018))\cite{03de2019deep}. PGnet with generation network are represents as PGnet W/G. PGnet with generation network and jump pattern are represents as PGnet W/G,J. PGnet with generation network, Jump pattern algorithm and motion evolution is represents as PGnet W/G,J,M.}
  \label{fig:comp}
\end{figure}
\subsection{Quantitative Results}
\paragraph{Ablation}
To better evaluate the effectiveness of our jump pattern algorithm and motion evolution method, we finally design a progressive ablation experiment.
Table \ref{tb:ablation} shows the experimental results of ablation on validation set. MSE, SSIM,PSNR are measured as average scores on every predicted frame. We can see that the generation network elevated the scores most. Jump pattern algorithm can improve MSE and PSNR benefits from the suppressed cumulative error by reducing the times of interpolation and deformation.
When it comes to motion evolution, the method can greatly increase the scores of all metrics because of the dynamic characteristics are learned by the proposed motion evolution method. The result is in line with our theoretical expectation and proves the effectiveness of the jump pattern and motion evolution.

\paragraph{Jump Pattern}
The MSE and SSIM scores of each step is shown in Figure \ref{fig:comp}. With the increase of the step, all model's prediction accuracy and similarity are decreasing. 
Our models outperform base model Emmanuel et al.(2019) \cite{03de2019deep} at every prediction steps.
and PGnet W/G,J,M achieves the best performance. 
The base model \cite{03de2019deep} seems not adapted to our instantaneous temperature dataset, which has more time correlations, and it can not generate pixels when conflict happens.
PGnet W/G,J produce a lower MSE score in the short term (first three steps) compared with PGnet W/G, but capture long term dynamics. The score gap between PGnet with and without the jump pattern algorithm indicates the effectiveness of the algorithm. 

\paragraph{Motion Evolution}

From Figure \ref{fig:comp} (a), the motion evolution method with momentum improves the MSE score by 0.43 on average for every prediction step. Figure \ref{fig:comp} (b) also indicates that the motion evolution method with momentum improves the SSIM at every step. The results show that our method can keep similarities and reduce errors by capturing the time-variant features of the motion field.
We also found that the motion evolution method leads to a faster convergence rate and a lower loss during training. 

\begin{figure}[htbp]
  \centering
  \begin{minipage}[t]{0.46\textwidth}
    \centering
    \parbox[][2.5cm][c]{\linewidth}{
     \hspace{0.35cm}
    \begin{subfigure}[t]{0.28\linewidth}
        \centering
            \includegraphics[width=1\linewidth]{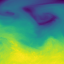}
        \caption{}
    \end{subfigure}
    \hspace{0.1cm}
    \begin{subfigure}[t]{0.28\linewidth}
        \centering
            \includegraphics[width=1\linewidth]{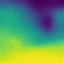}
        \caption{}
    \end{subfigure}
   \hspace{0.15cm}
    \begin{subfigure}[t]{0.28\linewidth}
        \centering
            \includegraphics[width=1\linewidth]{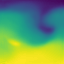}
         \caption{}
    \end{subfigure}}
    \caption{Comparison of the step-8 frame generated by models. (a) shows the ground truth, (b) generated by PGnet-Conv and (c) produced by PGnet-Momentum. The shape information of the vortex has been lost for PGnet-Conv but captured by PGnet-Momentum.}
  \label{fig:ME_comp}
  \end{minipage}
  \hspace{0.02\textwidth}
  \begin{minipage}[t]{0.5\textwidth}
    \centering
    \parbox[][2.5cm][c]{\linewidth}{
    \begin{subfigure}[t]{1\linewidth}
        \centering
        \includegraphics[width=1\linewidth]{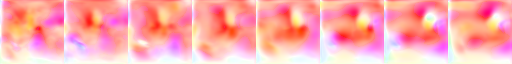}\vspace{0.08cm}
    \end{subfigure}
    \hspace{0.4cm}
    \begin{subfigure}[t]{1\linewidth}
        \centering
            \includegraphics[width=1\linewidth]{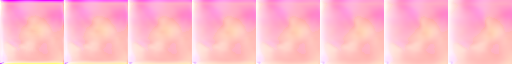}
    \end{subfigure}}
    \caption{A motion field sample of PGnet W/G (above images) and PGnet W/G,J,M (below images). Our model with motion evolution generates more consistent motion field results compared with the messy motion field generated by PGnet W/G .}
  \label{fig:comp_w}
  \end{minipage}
\end{figure}

In order to study the impact of different types of motion evolution methods, we compared the step-8 prediction frame of two variants of the motion evolution method mentioned in \ref{sec:me}. $\beta$ is set to $0.9999$ for PGnet-momentum that will be described below. From Figure \ref{fig:ME_comp}, the shape information of the vortex has been lost for PGnet-Conv, but captured by PGnet-Momentum even in the long term. PGnet-Momentum produces images that are closer to the real temperature field image.
One explanation is that time-varying features are more important than local constant spatial features for long-term prediction. 
As a result, PGnet-Conv with convolutional structure keeps more attention on spatial features and lost various detailed features of time in the long run. In contrast, PGnet-Momentum tracks the dynamic trajectory of every pixel and predicts a sharper image in the long term.
\begin{table}[h]
  \centering
  \caption{The table below shows average mean square error (MSE) score with different PGnet-Momentum coefficient values ($\beta$ in $F(.)$) :}\label{tab:score_beta}
  \begin{tabular}{ccccccc}
    \toprule  
    $\beta$ & 0 & 0.9 & 0.99&0.999&0.9999\\
    \midrule  
    MSE&9.40&9.42&9.16&8.98&8.87\\
    \bottomrule 
    \\
    \end{tabular}
\centering
\caption{Quantitative evaluation of different methods on the atmosphere temperature dataset. All metric scores are averaged by time steps. Higher PSNR , SSIM and CORR means better prediction accuracy. The coefficients for our model have been set by cross-validation with $\lambda_{lp} = 1$, $\lambda_{div} = 1$, $\lambda_{smoth} = 0.4$, $\alpha = 0.9$, $\beta = 0.9999$ for PGnet-Momentum.}
\label{tab:score_table}

\begin{tabular}{lllll}
  \toprule  
  Model& MSE& SSIM & PSNR& CORR\\
  \midrule  
  CDNN & 12.210& 0.820&28.884& 0.9724\\
  ConvLSTM& 9.121& 0.890&31.717& 0.9852\\
  DeepRNN& 9.287& 0.886 &31.657&0.9847\\
  PGnet-Conv&9.111&0.890&31.698& 0.9849\\
  PGnet-Momentum&\textbf{8.877}&\textbf{0.894}&\textbf{31.987}& \textbf{0.9860}\\
  \bottomrule 
  \end{tabular}
\end{table}
Besides, the impacts of the coefficient $\beta$ for method momentum is studied. From Table \ref{tab:score_beta}, When $\beta$ is in 0.999 $\sim$ 0.9999, it works reasonably fine. It indicates that a relatively large momentum is beneficial to motion. When $\beta$ is small (e.g., 0.9) or no momentum ($\beta$ is 0), MSE drops significantly; When $\beta$ is 1, there is no momentum, and the motion is consistent with the first motion field for every prediction step which is not considered in this paper. These results support our motivation of the motion field will not change dramatically. 
\paragraph{Evaluation}

Quantitatively, MSE, PSNR, SSIM scores of our model outperform other models. 
Table \ref{tab:score_table} shows the evaluation of our model 
and other baselines\cite{03de2019deep}\cite{25shi2017deep}\cite{pang2019deep}. 
Our model PGnet-Conv PGnet-momentum all outperforms other candidate models. And PGnet-momentum achieves the best performance on our dataset and respectively improve MSE scores by $2.6\%$ on average. 

Figure \ref{fig:comp_w} exhibits the motion field of our model. The picture above is predicted by PGnet W/G and the below picture is generated by PGnet W/G,J,M. As we see, PGnet without jump pattern is more messy and changeable. In contrast, the motion fields of different prediction steps with jump pattern and motion evolution are more similar and increasing with time steps and produce more convergent results.

\begin{figure*}[t]
  \centering
  \begin{subfigure}[t]{0.107\linewidth}
      \centering
      \makebox[0pt][r]{\makebox[14pt]{\raisebox{17pt}{\rotatebox[origin=c]{90}{\fontsize{6pt}{24pt} \textbf{GT}}}}}%
      \includegraphics[width=1.0\linewidth]{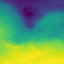}\vspace{0.08cm}
      \makebox[0pt][r]{\makebox[14pt]{\raisebox{17pt}{\rotatebox[origin=c]{90}{\fontsize{6pt}{24pt}   \textbf{CDNN}}}}}%
      \includegraphics[width=1.0\linewidth]{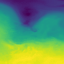}\vspace{0.08cm}
      \makebox[0pt][r]{\makebox[14pt]{\raisebox{17pt}{\rotatebox[origin=c]{90}{\fontsize{6pt}{24pt}  \textbf{ConvLstm}}}}}%
      \includegraphics[width=1.0\linewidth]{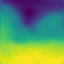}\vspace{0.08cm}
      \makebox[0pt][r]{\makebox[14pt]{\raisebox{17pt}{\rotatebox[origin=c]{90}{\fontsize{6pt}{24pt}  \textbf{DeepRnn}}}}}%
      \includegraphics[width=1.0\linewidth]{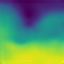}\vspace{0.08cm}
      \makebox[0pt][r]{\makebox[14pt]{\raisebox{17pt}{\rotatebox[origin=c]{90}{\fontsize{6pt}{24pt}  \textbf{PGnet}}}}}%
      \includegraphics[width=1.0\linewidth]{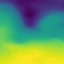}\vspace{0.08cm}
      \caption{step 1}
  \end{subfigure}\hspace{0.20cm}
  \begin{subfigure}[t]{0.107\linewidth}
      \centering
          \includegraphics[width=1.\linewidth]{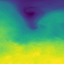}\vspace{0.08cm}
          \includegraphics[width=1.\linewidth]{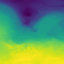}\vspace{0.08cm}
          \includegraphics[width=1.\linewidth]{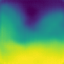}\vspace{0.08cm}
          \includegraphics[width=1.\linewidth]{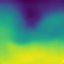}\vspace{0.08cm}
          \includegraphics[width=1.\linewidth]{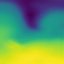}\vspace{0.08cm}
      \caption{step 2}
  \end{subfigure}\hspace{0.08cm}
  \begin{subfigure}[t]{0.107\linewidth}
      \centering
          \includegraphics[width=1.\linewidth]{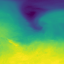}\vspace{0.08cm}
          \includegraphics[width=1.\linewidth]{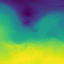}\vspace{0.08cm}
          \includegraphics[width=1.\linewidth]{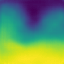}\vspace{0.08cm}
          \includegraphics[width=1.\linewidth]{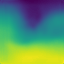}\vspace{0.08cm}
          \includegraphics[width=1.\linewidth]{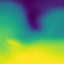}\vspace{0.08cm}
      \caption{step 3}
  \end{subfigure}\hspace{0.08cm}
  \begin{subfigure}[t]{0.107\linewidth}
      \centering
          \includegraphics[width=1.\linewidth]{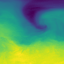}\vspace{0.08cm}
          \includegraphics[width=1.\linewidth]{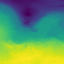}\vspace{0.08cm}
          \includegraphics[width=1.\linewidth]{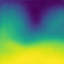}\vspace{0.08cm}
          \includegraphics[width=1.\linewidth]{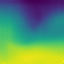}\vspace{0.08cm}
          \includegraphics[width=1.\linewidth]{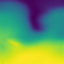}\vspace{0.08cm}
      \caption{step 4}
  \end{subfigure}\hspace{0.08cm}
  \begin{subfigure}[t]{0.107\linewidth}
      \centering
          \includegraphics[width=1.\linewidth]{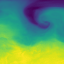}\vspace{0.08cm}
          \includegraphics[width=1.\linewidth]{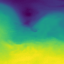}\vspace{0.08cm}
          \includegraphics[width=1.\linewidth]{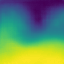}\vspace{0.08cm}
          \includegraphics[width=1.\linewidth]{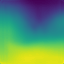}\vspace{0.08cm}
          \includegraphics[width=1.\linewidth]{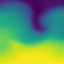}\vspace{0.08cm}
      \caption{step 5}
  \end{subfigure}\hspace{0.08cm}
  \begin{subfigure}[t]{0.107\linewidth}
      \centering
          \includegraphics[width=1.\linewidth]{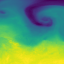}\vspace{0.08cm}
          \includegraphics[width=1.\linewidth]{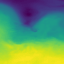}\vspace{0.08cm}
          \includegraphics[width=1.\linewidth]{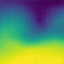}\vspace{0.08cm}
          \includegraphics[width=1.\linewidth]{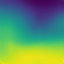}\vspace{0.08cm}
          \includegraphics[width=1.\linewidth]{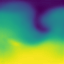}\vspace{0.08cm}
      \caption{step 6}
  \end{subfigure}\hspace{0.08cm}
  \begin{subfigure}[t]{0.107\linewidth}
      \centering
          \includegraphics[width=1.\linewidth]{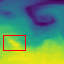}\vspace{0.08cm}
          \includegraphics[width=1.\linewidth]{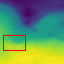}\vspace{0.08cm}
          \includegraphics[width=1.\linewidth]{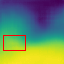}\vspace{0.08cm}
          \includegraphics[width=1.\linewidth]{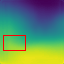}\vspace{0.08cm}
          \includegraphics[width=1.\linewidth]{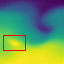}\vspace{0.08cm}
      \caption{step 7}
  \end{subfigure}\hspace{0.08cm}
  \begin{subfigure}[t]{0.107\linewidth}
      \centering
          \includegraphics[width=1.\linewidth]{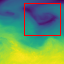}\vspace{0.08cm}
          \includegraphics[width=1.\linewidth]{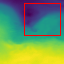}\vspace{0.08cm}
          \includegraphics[width=1.\linewidth]{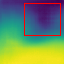}\vspace{0.08cm}
          \includegraphics[width=1.\linewidth]{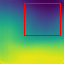}\vspace{0.08cm}
          \includegraphics[width=1.\linewidth]{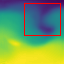}\vspace{0.08cm}
      \caption{step 8}
  \end{subfigure}
  \caption{
  Experiment results of different models in the 2 days (8 steps) temperature field image prediction task. The time difference between each consecutive step is 6 hours. The bright spot shown in the small red box indicates the highest temperature, while the vortex in the large red box represents dramatic temperature changes. Our new model captured bright spot and vortex, which are distinct from the absence of these features in other models.
 }
\vspace{-3mm}
\label{fig:Nowcastingrst}
\end{figure*}

Figure \ref{fig:Nowcastingrst} presents the ground truth and the predicted frames of different models. The PGnet model using momentum-based $F(.)$ with $\beta = 0.9999$ and evolution order $M=1$. All models are given 4 frames as input and required to predict the next 8 frames. In contrast to the base model CDNN with regularization\cite{03de2019deep}, other methods capture more dynamics. It indicates that although the CDNN model is constrained by physical equations and produces clear prediction pictures, it can not learn the dynamics. ConvLSTM and DeepRNN model seems to capture some dynamics but suffers more from blurriness with the step increases because of the lack of physical constraints and cumulative errors. Besides, our model benefits from both physics and deep learning methods. It learns higher-order dynamics by introducing the motion evolution method. As demonstrated in Figure \ref{fig:Nowcastingrst} that both the bright spot and vortex shown in red boxes are captured by our model, while other models are missed. PGnet predictions are the closest to ground truth and captures more details that are essential to weather forecasting.


\section{Conclusion}
\label{sec:conclusion}
In this study, we introduce PGnet that incorporates a physic-informed method and an image synthesis neural network for troposphere temperature prediction. Towards learning time-variant features and increase accuracy, the motion field evolution method with a jump pattern strategy are proposed. We validate our proposal through a series of ablation experiments and provide exhaustive comparisons of PGnet with baselines. Quantitative studies on the ERA5  500 hPa level  temperature dataset demonstrate that the physics-informed model can effectively improve the accuracy of upper atmosphere temperature prediction. The proposed method's effectiveness on other variables and levels is an exciting research area that we will study in the future.


\renewcommand*{\thesection}{\the\value{section}}

    
\bibliographystyle{iopart-num}  
\bibliography{egbib_pp}

\providecommand{\newblock}{}
\begin{thebibliography}{10}
\expandafter\ifx\csname url\endcsname\relax
  \def\url#1{{\tt #1}}\fi
\expandafter\ifx\csname urlprefix\endcsname\relax\def\urlprefix{URL }\fi
\providecommand{\eprint}[2][]{\url{#2}}

\bibitem{Kornhuber-ERL}
Kornhuber K, Osprey S, Coumou D, Petri S, Petoukhov V, Rahmstorf S and Gray L
  2019 {\em Envrionmental Research Letters\/} {\bf 14}

\bibitem{Thornton-ERL}
Thornton H, Scaife A, Hoskins B~J and Brayshaw D 2017 {\em Envrionmental
  Research Letters\/} {\bf 12}

\bibitem{02lynch2008origins}
Lynch P 2008 {\em Journal of Computational Physics\/} {\bf 227} 3431--3444

\bibitem{stanger2019optimising}
Stanger J, Finney I, Weisheimer A and Palmer T 2019 {\em Environmental Research
  Letters\/} {\bf 14} 124086

\bibitem{01sekula2019prediction}
Sekula P, Bokwa A, Bochenek B and Zimnoch M 2019 {\em Atmosphere\/} {\bf 10}
  186

\bibitem{30brunton2016discovering}
Brunton S~L, Proctor J~L and Kutz J~N 2016 {\em Proceedings of the national
  academy of sciences\/} {\bf 113} 3932--3937

\bibitem{10ham2019deep}
Ham Y~G, Kim J~H and Luo J~J 2019 {\em Nature\/} {\bf 573} 568--572

\bibitem{09roscher2020explainable}
Roscher R, Bohn B, Duarte M~F and Garcke J 2020 {\em IEEE Access\/}

\bibitem{Yan2020}
Wang M, Yan Z, Lu J and Chen X 2020 {\em Nature Electronics\/} {\bf 3}

\bibitem{Feng2021}
Feng Z, Niu W, Tang Z, Xu Y and Zhang H 2021 {\em Journal of Hydrology\/} {\bf
  595} 126062

\bibitem{Feng2021b}
Feng Z and Niu W 2021 {\em Knowledge-Based Systems\/} {\bf 211} 106580

\bibitem{24xingjian2015convolutional}
Xingjian S, Chen Z, Wang H, Yeung D~Y, Wong W~K and Woo W~c 2015 Convolutional
  lstm network: A machine learning approach for precipitation nowcasting {\em
  Advances in neural information processing systems\/} pp 802--810

\bibitem{wang2018predrnn++}
Wang Y, Gao Z, Long M, Wang J and Philip S~Y 2018 Predrnn++: Towards a
  resolution of the deep-in-time dilemma in spatiotemporal predictive learning
  {\em International Conference on Machine Learning\/} (PMLR) pp 5123--5132

\bibitem{jin2020exploring}
Jin B, Hu Y, Tang Q, Niu J, Shi Z, Han Y and Li X 2020 Exploring
  spatial-temporal multi-frequency analysis for high-fidelity and
  temporal-consistency video prediction {\em Proceedings of the IEEE/CVF
  Conference on Computer Vision and Pattern Recognition\/} pp 4554--4563

\bibitem{25shi2017deep}
Shi X, Gao Z, Lausen L, Wang H, Yeung D~Y, Wong W~k and Woo W~c 2017 Deep
  learning for precipitation nowcasting: A benchmark and a new model {\em
  Advances in neural information processing systems\/} pp 5617--5627

\bibitem{pang2019deep}
Pang B, Zha K, Cao H, Shi C and Lu C 2019 Deep rnn framework for visual
  sequential applications {\em Proceedings of the IEEE Conference on Computer
  Vision and Pattern Recognition\/} pp 423--432

\bibitem{lstmgreff2016lstm}
Greff K, Srivastava R~K, Koutn{\'\i}k J, Steunebrink B~R and Schmidhuber J 2016
  {\em IEEE transactions on neural networks and learning systems\/} {\bf 28}
  2222--2232

\bibitem{liang2017dual}
Liang X, Lee L, Dai W and Xing E~P 2017 Dual motion gan for future-flow
  embedded video prediction {\em Proceedings of the IEEE International
  Conference on Computer Vision\/} pp 1744--1752

\bibitem{38wang2018video}
Wang T~C, Liu M~Y, Zhu J~Y, Liu G, Tao A, Kautz J and Catanzaro B 2018
  Video-to-video synthesis {\em Conference on Neural Information Processing
  Systems (NeurIPS)\/}

\bibitem{ilg2017flownet}
Ilg E, Mayer N, Saikia T, Keuper M, Dosovitskiy A and Brox T 2017 Flownet 2.0:
  Evolution of optical flow estimation with deep networks {\em Proceedings of
  the IEEE conference on computer vision and pattern recognition\/} pp
  2462--2470

\bibitem{liu2017video}
Liu Z, Yeh R~A, Tang X, Liu Y and Agarwala A 2017 Video frame synthesis using
  deep voxel flow {\em Proceedings of the IEEE International Conference on
  Computer Vision\/} pp 4463--4471

\bibitem{kwon2019predicting}
Kwon Y~H and Park M~G 2019 Predicting future frames using retrospective cycle
  gan {\em Proceedings of the IEEE Conference on Computer Vision and Pattern
  Recognition\/} pp 1811--1820

\bibitem{goodfellow2014generative}
Goodfellow I, Pouget-Abadie J, Mirza M, Xu B, Warde-Farley D, Ozair S,
  Courville A and Bengio Y 2014 Generative adversarial nets {\em Advances in
  neural information processing systems\/} pp 2672--2680

\bibitem{zhu2017unpaired}
Zhu J~Y, Park T, Isola P and Efros A~A 2017 Unpaired image-to-image translation
  using cycle-consistent adversarial networks {\em Proceedings of the IEEE
  international conference on computer vision\/} pp 2223--2232

\bibitem{han2018solving}
Han J, Jentzen A and Weinan E 2018 {\em Proceedings of the National Academy of
  Sciences\/} {\bf 115} 8505--8510

\bibitem{03de2019deep}
de~Bezenac E, Pajot A and Gallinari P 2019 {\em Journal of Statistical
  Mechanics: Theory and Experiment\/} {\bf 2019} 124009

\bibitem{06onfared2009new}
Monfared M, Rastegar H and Kojabadi H~M 2009 {\em Renewable energy\/} {\bf 34}
  845--848

\bibitem{raissi2019physics}
Raissi M, Perdikaris P and Karniadakis G~E 2019 {\em Journal of Computational
  Physics\/} {\bf 378} 686--707

\bibitem{31berg2019data}
Berg J and Nystr{\"o}m K 2019 {\em Journal of Computational Physics\/} {\bf
  384} 239--252

\bibitem{32raissi2018deep}
Raissi M 2018 {\em The Journal of Machine Learning Research\/} {\bf 19}
  932--955

\bibitem{33seo2019differentiable}
Seo S and Liu Y 2019 {\em arXiv preprint arXiv:1902.02950\/}

\bibitem{seo2019physics}
Seo S, Meng C and Liu Y 2019 Physics-aware difference graph networks for
  sparsely-observed dynamics {\em International Conference on Learning
  Representations\/}

\bibitem{wang2020towards}
Wang R, Kashinath K, Mustafa M, Albert A and Yu R 2020 Towards physics-informed
  deep learning for turbulent flow prediction {\em Proceedings of the 26th ACM
  SIGKDD International Conference on Knowledge Discovery \& Data Mining\/} pp
  1457--1466

\bibitem{34weinan2017proposal}
Weinan E 2017 {\em Communications in Mathematics and Statistics\/} {\bf 5}
  1--11

\bibitem{long2018pde}
Long Z, Lu Y, Ma X and Dong B 2018 Pde-net: Learning pdes from data {\em
  International Conference on Machine Learning\/} (PMLR) pp 3208--3216

\bibitem{36long2019pde}
Long Z, Lu Y and Dong B 2019 {\em Journal of Computational Physics\/} {\bf 399}
  108925

\bibitem{37guen2020disentangling}
Guen V~L and Thome N 2020 Disentangling physical dynamics from unknown factors
  for unsupervised video prediction {\em Proceedings of the IEEE/CVF Conference
  on Computer Vision and Pattern Recognition\/} pp 11474--11484

\bibitem{27gao2019disentangling}
Gao H, Xu H, Cai Q~Z, Wang R, Yu F and Darrell T 2019 Disentangling propagation
  and generation for video prediction {\em Proceedings of the IEEE
  International Conference on Computer Vision\/} pp 9006--9015

\bibitem{bertalmio2000image}
Bertalmio M, Sapiro G, Caselles V and Ballester C 2000 Image inpainting {\em
  Proceedings of the 27th annual conference on Computer graphics and
  interactive techniques\/} pp 417--424

\bibitem{barnes2009patchmatch}
Barnes C, Shechtman E, Finkelstein A and Goldman D~B 2009 {\em ACM Trans.
  Graph.\/} {\bf 28} 24

\bibitem{zhang2018context}
Zhang H, Dana K, Shi J, Zhang Z, Wang X, Tyagi A and Agrawal A 2018 Context
  encoding for semantic segmentation {\em Proceedings of the IEEE conference on
  Computer Vision and Pattern Recognition\/} pp 7151--7160

\bibitem{pathak2016context}
Pathak D, Krahenbuhl P, Donahue J, Darrell T and Efros A~A 2016 Context
  encoders: Feature learning by inpainting {\em Proceedings of the IEEE
  conference on computer vision and pattern recognition\/} pp 2536--2544

\bibitem{mcnetvillegas2017decomposing}
Villegas R, Yang J, Hong S, Lin X and Lee H 2017 Decomposing motion and content
  for natural video sequence prediction {\em 5th International Conference on
  Learning Representations, ICLR 2017\/} (International Conference on Learning
  Representations, ICLR)

\bibitem{nemgreff2017neural}
Greff K, Van~Steenkiste S and Schmidhuber J 2017 Neural expectation
  maximization {\em Advances in Neural Information Processing Systems\/} pp
  6691--6701

\bibitem{zhou2020deep}
Zhou Y, Dong H and El~Saddik A 2020 {\em IEEE Access\/} {\bf 8} 69273--69283

\bibitem{cramer1961some}
Cram{\'e}r H 1961 On some classes of nonstationary stochastic processes {\em
  Proceedings of the Fourth Berkeley symposium on mathematical statistics and
  probability\/} vol~2 (University of Los Angeles Press Berkeley and Los
  Angeles) pp 57--78

\bibitem{lucas1981iterative}
Lucas B~D, Kanade T {\em et~al.\/} 1981

\bibitem{finn2016unsupervised}
Finn C, Goodfellow I and Levine S 2016 Unsupervised learning for physical
  interaction through video prediction {\em Advances in neural information
  processing systems\/} pp 64--72

\bibitem{hao2018controllable}
Hao Z, Huang X and Belongie S 2018 Controllable video generation with sparse
  trajectories {\em Proceedings of the IEEE Conference on Computer Vision and
  Pattern Recognition\/} pp 7854--7863

\bibitem{hersbach2020era5}
Hersbach H, Bell B, Berrisford P, Hirahara S, Hor{\'a}nyi A, Mu{\~n}oz-Sabater
  J, Nicolas J, Peubey C, Radu R, Schepers D {\em et~al.\/} 2020 {\em Quarterly
  Journal of the Royal Meteorological Society\/} {\bf 146} 1999--2049

\bibitem{molteni1996ecmwf}
Molteni F, Buizza R, Palmer T~N and Petroliagis T 1996 {\em Quarterly journal
  of the royal meteorological society\/} {\bf 122} 73--119

\bibitem{Zhang_2018_CVPR}
Zhang X, Zhou X, Lin M and Sun J 2018 Shufflenet: An extremely efficient
  convolutional neural network for mobile devices {\em Proceedings of the IEEE
  Conference on Computer Vision and Pattern Recognition (CVPR)\/}

\end{thebibliography}

\end{document}